\documentclass[runningheads]{llncs}

\usepackage[T1]{fontenc}
\usepackage{graphicx}
\usepackage{amsmath}
\usepackage{amsfonts}
\usepackage{booktabs}
\usepackage{units}
\usepackage{xspace}
\usepackage{hyperref}

\begin{document}

\newcommand{\AN}{\textbf{AN}\xspace}
\newcommand{\ANT}{\textbf{AN$_\mathbf{t}$}\xspace}
\newcommand{\ANTk}{\textbf{AN$_\mathbf{t}^\mathbf{k}$}\xspace}

\title{Stochastic Filtering for Quorum Sensing in Robot Swarms under Anonymous Communication}

\titlerunning{Quorum Sensing in Robot Swarms under Anonymous Communication}

\author{Fabio Oddi\inst{1,2}\orcidID{0009-0001-8379-0209} \and Andreagiovanni Reina\inst{3,4,5}\orcidID{0000-0003-4745-992X} \and Vito Trianni\inst{2}\orcidID{0000-0002-9114-8486}}

\authorrunning{F. Oddi et al.}

\institute{DIAG, Sapienza University of Rome, Italy \and ISTC, National Research Council, Rome, Italy \and Centre for the Advanced Study of Collective Behaviour, University of Konstanz, Konstanz, Germany \and Department of Computer and Information Science, University of Konstanz, Konstanz, Germany \and 
Max Planck Institute of Animal Behavior, Konstanz, Germany \\
\email{oddi@diag.uniroma1.it, andreagiovanni.reina@uni-konstanz.de, vito.trianni@istc.cnr.it}}

\index{Oddi, Fabio}
\index{Reina, Andreagiovanni}
\index{Trianni, Vito}

\maketitle
\setcounter{footnote}{0}

\begin{abstract}
Quorum Sensing (QS) is a key capability for robot swarms, useful for coordination of activities at the group level. Effective communication is instrumental for individuals to estimate the quorum level of the entire swarm. Anonymous communication protocols where individuals exchange local information without revealing unique identities are helpful to support quorum estimates by sampling information from neighbours and maintain scalability of the QS process. However, because anonymous protocols cannot distinguish message sources, repeated messages from the same sender may be double-counted, thereby biasing collective quorum estimates. In this study, we introduce a stochastic filtering protocol inspired by $k$-priority sampling to improve estimate stability (\ANTk), and we compare it with a baseline anonymous protocols (\AN) and a randomised variant designed to improve accuracy (\ANT). We find that the baseline protocol \AN provides a parsimonious and fast solution, but remains highly inaccurate due to double-counting bias. The \ANT variant improves accuracy but suffers from information inertia, resulting in slower convergence. Finally, actively filtering the message buffer via the \ANTk protocol successfully decreases temporary errors and stabilises the estimate, at the cost of an increased time of recovery from errors.
\end{abstract}


\section{Introduction}\label{sec:intro}
In decentralised systems, global coordination emerges from local, anonymous interactions, a phenomenon widely observed in biological collectives. Natural populations, such as bacterial colonies and social insects, rely on quorum sensing (QS) to execute group-level tasks. For instance, bacteria produce and accumulate diffusible autoinducer molecules to assess their surroundings, which subsequently triggers synchronised shifts in gene expression~\cite{miller2001quorum,cornforth2014combinatorial}. Chemical communication is crucial also for social insects like ants, which also combine pheromones with vibro-acoustic stridulation to enhance the precision of their communications~\cite{ciaralli2026multimodal,golden2016evolution}. Furthermore, species such as \textit{Temnothorax albipennis} rely on the rate of direct tactile encounters with nest-mates to gauge local population density, employing a physical QS mechanism during nest relocation~\cite{pratt2005quorum}. By contrast, honeybees employ modulatory vibrational signals to synchronise labour and the selection of nest sites~\cite{schneider2004vibration}.

Drawing inspiration from biological mechanisms, swarm robotics systems have utilised QS to support ``best-of-$n$'' decision making, where a swarm must collectively choose the highest-quality option from a set of alternatives~\cite{martinez2020bacterial,oddi2022best-of-n-005}. To successfully reach a consensus and prevent decision deadlocks, swarms employ strategies such as cross-inhibition~\cite{zakir2022robot} and interaction modulation~\cite{talamali2019improving}. Recent research also suggests that the introduction of measurement errors or miscommunication amongst robots can occasionally enhance the overall accuracy of the group during such decisions~\cite{zakir2024miscommunication}. Crucially, the efficacy of these high-level consensus strategies to drive coordinated action fundamentally depends on a robust underlying protocol for QS. Indeed, without an accurate and robust mechanism for QS, individual robots within the swarm cannot detect the achievement of consensus by themselves, and therefore cannot act accordingly. If QS is not accurate, the swarm faces the risk of splitting and coordination may get lost.

When deploying QS onto physical robots, achieving accurate QS presents a practical challenge. To ensure scalability, minimalist approaches require to operate with anonymous communication protocols, hence avoiding to share information about the sender identity and therefore also minimising the amount of information to store.\footnote{See~\cite{Oddi2025sub} for a wide comparison between anonymous and identity-aware protocols.} Yet, in the context of QS, the absence of unique identifiers inherently leads to the double-counting of opinions from neighbours~\cite{Oddi2024}, resulting in estimation biases. More precisely, if the robot interaction network does not change frequently, it is likely that multiple messages broadcast by some robot are received and stored as if they were different messages, due to their anonymity. Any protocol that systematically uses duplicate messages will therefore suffer from a loss in accuracy due to ineffective sampling. To address message redundancy without unique identifiers, distributed systems often draw from distinct element estimation and probabilistic counting streaming frameworks~\cite{bar2002counting,flajolet2007hyperloglog}.

In previous work, we introduced a baseline protocol (\AN) and its accuracy-enhancing variant (\ANT)~\cite{Oddi2025sub}. The baseline \AN is based on a FIFO buffer that gets updated every time a new message is received. By simply reshuffling this buffer, \ANT demonstrates improved accuracy at the cost of retaining old information, resulting in slower updates.  
Here, we introduce the \ANTk protocol to mitigate the estimation instability arising from double-counting. Adapting the concept of $k$-priority sampling---initially formulated to estimate subset sums within data streams~\cite{duffield2007priority,daliri2024simple,thorup2013bottom}---this protocol modulates the adverse effects of local data redundancy. By neglecting a portion of the messages that are closest to expiration, the \ANTk protocol operates on the assumption that the discarded data points are likely redundant and remain represented within the buffer by more recent duplicate messages.

The paper is organised as follows: Section~\ref{sec:qstc} formalises the QS process along with the anonymous protocols and the proposed stochastic filtering. Section~\ref{sec:results} outlines the experimental methodology and analyses the simulation results, evaluating the protocols across multiple metrics. Finally, Section~\ref{sec:conclusions} summarises the main findings and discusses the fundamental trade-offs identified in the study.

\section{Quorum Sensing and Stochastic Filtering}\label{sec:qstc}
We consider a swarm of $N$ robots operating under anonymous communication, meaning messages are collected with no sender identity. A fraction $G \in [0,1]$ of the swarm---referred to as the \textit{ground truth}---is initialised in the \textit{committed} state. We compute the fraction of robots $Q \in [0,1]$ that recognise when the percentage of the swarm being committed surpasses a given \textit{threshold} $\tau \in [0,1]$. Ideally, as shown in previous work~\cite{Oddi2024,Oddi2025sub}, perfect QS corresponds to $Q = 1 \iff G \geq \tau$ and $Q = 0 \iff G < \tau$.

To estimate the ground truth $G$, robots broadcast messages containing their state to neighbours within a radius $r$ while moving across the working area. Each robot $i$ stores received messages in a buffer $\mathcal{B}_i$ of maximum capacity $B_M$.
After each update of $\mathcal{B}_i$, the robot evaluates the fraction of messages received from \textit{committed} robots against the threshold $\tau$. To ensure a sufficient sample size, we impose a minimum buffer size $B_m$. The individual robot's one-bit quorum detection state $b_q(i)$ and the global population estimation $Q$ are computed at each iteration as follows:
\begin{align}
    b_q(i) &= |\mathcal{B}_i|\geq B_m \wedge \sum_{m\in\mathcal{B}_i}b_c(m) \geq \tau|\mathcal{B}_i|,\\
    Q &=\frac{1}{N}\sum_{i=1}^N b_q(i),
\end{align}
where $b_c(m)\in\{0,1\}$ is the commitment state stored in message $m$. The variable $b_q(i)$ fluctuates between $0$ and $1$ depending on the messages stored in $\mathcal{B}_i$. When available information is insufficient ($|\mathcal{B}_i| < B_m$), the quorum is considered undetected ($b_q(i)=0$).

\subsection{Baseline Anonymous Protocols}\label{subsubsec:anonym_prot}
To implement the baseline anonymous protocol \AN, the message $m$ must only contain the robot state $b_c(m)$. Received messages are managed via a standard FIFO logic. Therefore, the maximum memory required by this protocol is $B_M$ bits.
A variant protocol (\ANT) is designed to improve estimation accuracy by reordering the messages stored in $\mathcal{B}_i$. To this end, upon reception of a message $m$, a random timeout $t(m)$ is drawn from an exponential distribution with average $T_m$. These timeouts are updated over time for messages stored in the buffer, and govern the insertion of newly received messages so that the buffer remains sorted in descending order of timeout. In this protocol, the information stored for each message is the tuple $\langle b_c(m),t(m) \rangle$. When the buffer fills and a new message arrives, rather than discarding the oldest entry, the protocol removes the message with the lowest remaining timeout. Because timeouts are drawn randomly from an exponential distribution, message insertion practically corresponds to random shuffling of the buffer, which breaks the temporal correlation inherent in receiving continuous, redundant broadcasts from the same nearby neighbour. By scrambling message retention, the \ANT protocol enables the buffer to maintain greater source diversity, effectively gathering a wider spatial sample. By updating timeouts as time passes, the \ANT protocol can simulate different buffer behaviours: from a normal FIFO (when $T_m$ is much smaller than the insertion rate) to random insertion (when $T_m$ is large). Using one bit for the commitment state and $B_t$ bits for timeouts, this protocol requires at most $B_M(B_t+1)$ bits.

\subsection{Stochastic Filtering}\label{subsubsec:k_priority_prot}
To address the stability limitations observed in the previous protocols without compromising anonymity, we introduce a second variant protocol (\ANTk) designed specifically to improve estimate stability. While \ANT leverages timeouts solely for buffer replacement, the \ANTk protocol actively integrates these temporal values into the estimation logic itself. 

When the buffer $\mathcal{B}_i$ reaches its maximum capacity $B_M$, this protocol treats the random message timeouts $t(m)$ as stochastic weights. Specifically, while computing the quorum fraction, the $k$ elements with the lowest residual timeout are systematically neglected from the calculation. Crucially, these elements are not deleted from the memory; they are merely masked during the current evaluation cycle. The underlying assumption is that messages closest to expiration are statistically more likely to be redundant duplicates of information already represented in the buffer by messages with longer timeouts. By continuously filtering out this expiring tail of data, the \ANTk protocol acts as a low-pass filter on the incoming information stream. It systematically smooths out the high-frequency fluctuations caused by double-counting, trading a slight reduction in the effective sample size for enhanced consensus stability.

\section{Analysis of Experimental Results}
\label{sec:results}
We conducted simulations within the ARGoS framework~\cite{Pinciroli2012dc,Pinciroli2018ants}, employing the Kilobot~\cite{Rubenstein2014dq} as our reference swarm platform. These robots operate under severe hardware constraints, featuring a maximum velocity of $\unit[1]{cm/s}$, a message transmission range $r=\unit[10]{cm}$, a frequency bounded at $\unit[2]{Hz}$, and a memory capacity of one kilobyte. Kilobots move within the arena following a random waypoint model, exploiting the Augmented Reality for Kilobots (ARK) system~\cite{Reina2017ef,Pinciroli2018ants} to provide absolute positioning information.
More specifically, knowing the arena boundaries, the robot uniformly samples a random target location and proceeds towards it in a straight line.
A new destination is generated whenever an agent arrives at its target or becomes physically stuck. To identify stationary stalling, the protocol imposes a temporal limit on every journey: robots are allotted the theoretical time required to traverse the distance at maximum speed, supplemented by a 33\% tolerance margin. 
Alongside this passive timeout mechanism, robots utilise a proactive collision-resolution strategy powered by the ARK system. By monitoring their position, robots compare the current distance to the destination against the smallest value reached so far. If this distance increases by $\unit[1]{cm}$ or more, the robot infers it is being pushed backwards by peers. Upon detecting this active displacement, the robot engages a stochastic evasive manoeuvre, randomly electing to steer left, steer right, or forge straight ahead until it registers a clear reduction in its distance to the goal. After disengagement, the robot recalculates a trajectory to its original target waypoint.

Robots are initially randomly placed in a square arena and assigned a static commitment state they never change: $\lceil GN\rceil$ robots are initialised as committed, while $\lfloor(1-G)N\rfloor$ robots are uncommitted. This static commitment choice allows us to focus exclusively on the raw quorum sensing and estimation process, decoupling it from the consensus-building dynamics.
We tested 13 values for the ground truth $G$ ranging from $0.52$ to $1$ with a step of $0.04$, while the threshold $\tau$ varied between $0.5$ and $1$ with steps of $0.01$. We study three deployment scenarios---HD25, LD25 and HD100---that vary both swarm size and arena size. In the HD25 case, a small swarm of 25 robots operates in a high-density scenario being placed in a small arena of $\unit[0.25]{m^2}$. In the LD25 case, the same 25-robot swarm operates in low-density scenarios, being placed in a large arena of $\unit[1]{m^2}$. Finally, in the HD100 case, a larger swarm of 100 robots operates in high-density scenarios (same as for HD25) being placed in a large arena of $\unit[1]{m^2}$.
All anonymous protocols use a buffer size $B_M=N-1$ and a minimum number of messages $B_m=5$. To evaluate the impact of the stochastic filtering, we investigated three values for the parameter $k$ within the \ANTk protocol. Specifically, we examined the system's performance across $k_1 = 1$, $k_2=\frac{N}{5}$, $k_3=\frac{2N}{5}$.

\subsection{Message Diversity}
\begin{figure}[!t]
    \centering
    \includegraphics[width=1\textwidth]{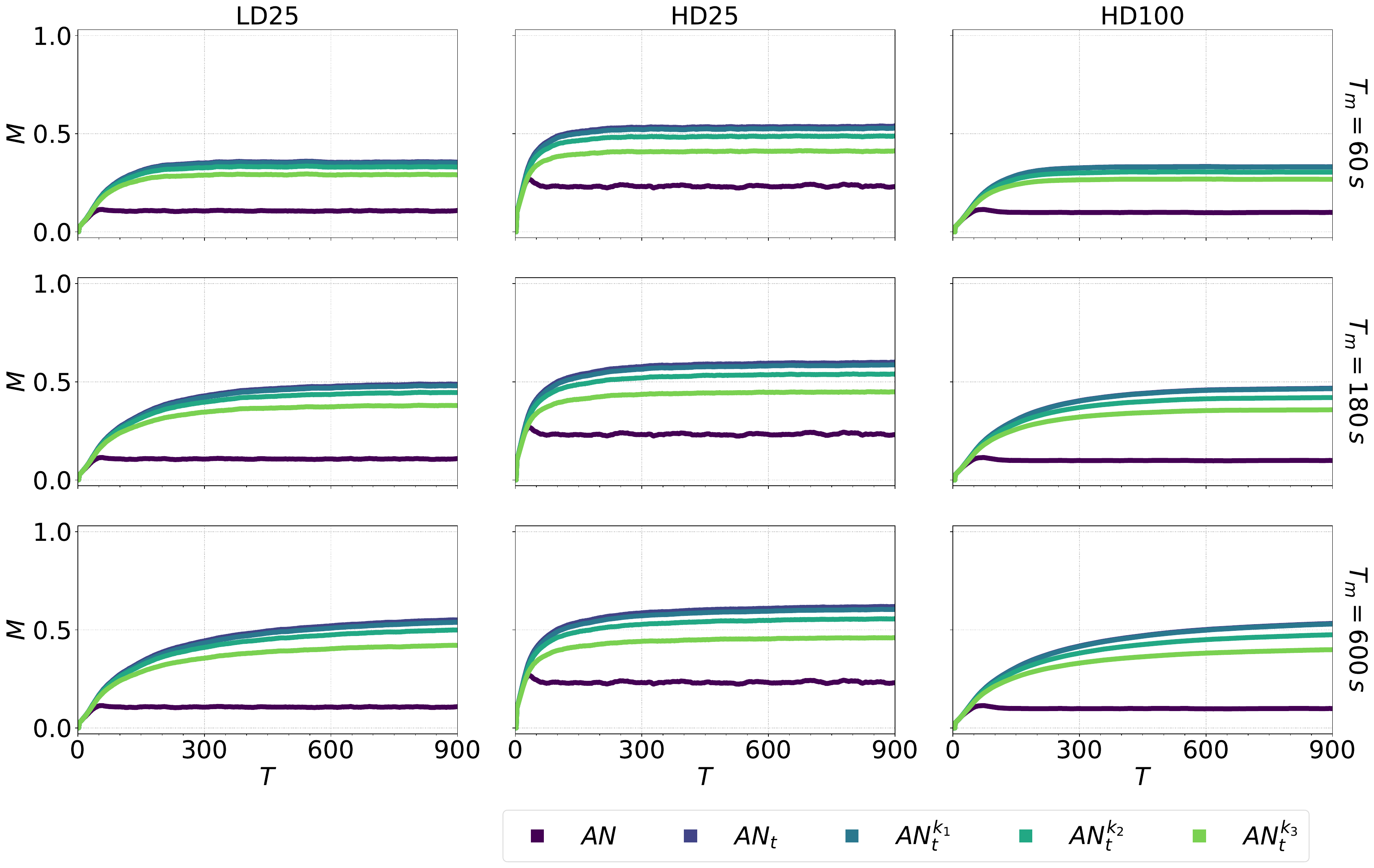}
    \caption{Evolution over time of the average message diversity $M$. The different scenarios (LD25, HD25, HD100) are sorted by columns from left to right. Different timeout values ($T_m$) are sorted by rows in ascending order.}
    \label{fig:messages}
\end{figure}

We first study the fraction $M$ of unique messages in the buffer, displayed in Figure~\ref{fig:messages}, which serves as an indicator of the useful information available to each robot. Higher message diversity generally corresponds to better information regarding the swarm, yielding higher QS accuracy. Note that robots have no mean to evaluate $M$, but this measure is useful to evaluate the ability of the protocols to collect and retain information from different sources.

Since the baseline protocol \AN lacks a timeout mechanism and relies on a strict FIFO replacement policy, its buffer is rapidly saturated by redundant messages from immediate neighbours. Consequently, the number of unique messages $M$ remains small across different densities and is unaffected by $T_m$. The accuracy variant protocol \ANT mitigates this loss of information by shuffling the buffer. High values of $T_m$ allow it to preserve older, unique messages, yielding improved diversity compared to \AN. In the short timeout regime ($T_m = 60$\,s), \ANT exhibits a more rapid message turnover, which reduces diversity, though it retains more unique messages than the baseline \AN. By filtering the lowest-priority data, the stability variant protocol \ANTk maintains a lower fraction of unique identifiers within the buffer compared to \ANT (especially for $k_2$ and $k_3$).

By looking at the different density scenarios, we note that message diversity is mainly determined by the arena size rather than by the density. For all protocols, $M$ is higher is the high-density, small-arena scenario HD25, meaning that robots communicate and mix more effectively thanks to the reduced distances to be covered. On the contrary, diversity is low in both LD25 and HD100 scenarios, due to the dominance of local interactions.

\subsection{QS Accuracy}
\begin{figure}[!t]
    \centering
    \includegraphics[width=1\textwidth]{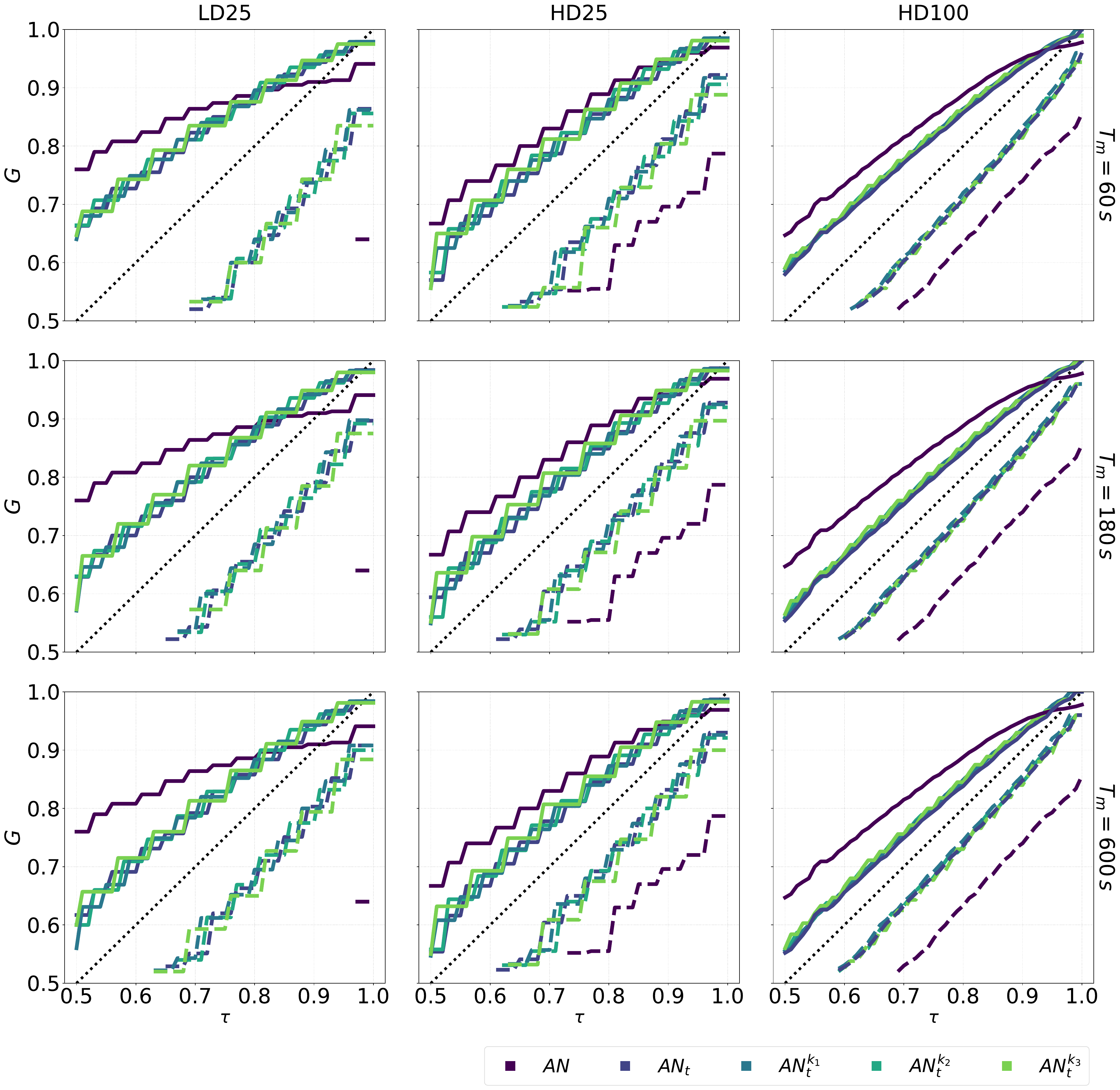}
    \caption{Isolines of the average quorum detection for $Q=0.8$ (solid lines) and $Q=0.2$ (dashed lines). The different scenarios (LD25, HD25, HD100) are sorted by columns from left to right. Different timeout values ($T_m$) are sorted by rows in ascending order.}
    \label{fig:isolines}
\end{figure}

To determine the accuracy of the QS process, we consider two decision thresholds. When at least a fraction $Q=0.8$ of the swarm recognises that quorum is achieved, we flag a detection. When at most a fraction $Q=0.2$ of the swarm recognises that the quorum is achieved, we flag a rejection.  In Figure~\ref{fig:isolines} we plot the boundaries of these regions as isolines: solid for $Q=0.8$, dashed for $Q=0.2$. The proximity of these boundaries to the ideal dividing line $G=\tau$ defines overall accuracy, because the region in which the swarm is undecided is minimised. 

The baseline protocol \AN demonstrates a systemic bias across all density scenarios, requiring a committed majority much larger than the nominal threshold to trigger detection reliably. Additionally, when $G>0.9$, this protocols suffers from false positives, where detection is observed for conditions in which $G<\tau$. This inaccuracy results from low message diversity and highly redundant local information. The variant protocol \ANT generally narrows the uncertainty gap and limits the occurrence of false positives, which are observed only for small swarm sizes (LD25 and HD25).
Notably, the \ANTk protocol does not solve the double-counting problem due to spatial correlations. Indeed, the constraints of identifier-free communication impose that spatial correlation cannot be entirely eliminated. Still, we note that there is no significant loss of accuracy with respect to \ANT despite the lower messages diversity $M$ observed in Figure~\ref{fig:messages}. 

\subsection{QS Latency}
\begin{figure}[!b]
    \centering
    \includegraphics[width=1\textwidth]{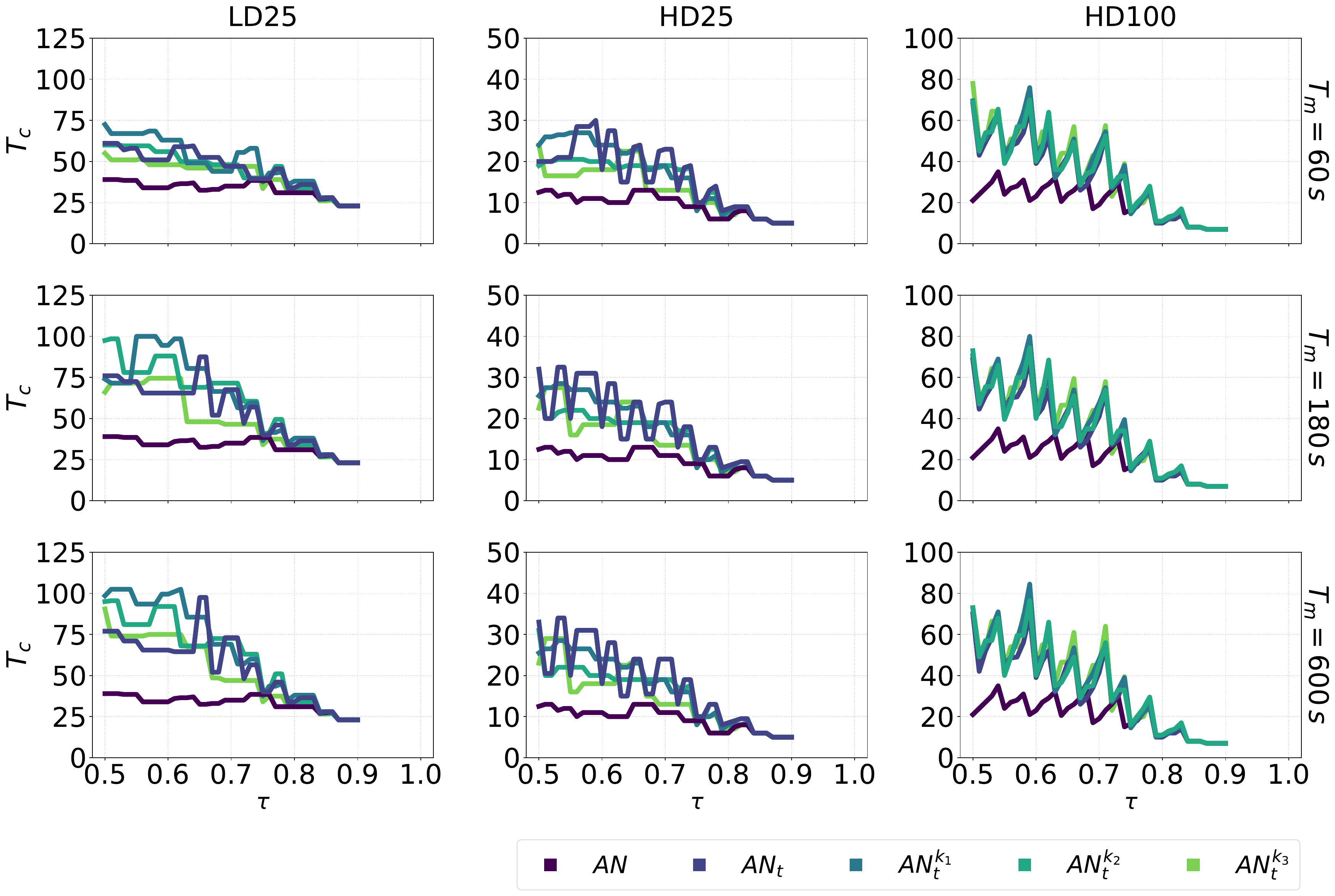}
    \caption{Median recognition latency, computed as the time needed for the swarm to reliably detect the quorum ($Q\geq0.8$). Data are the average over $R=100$ simulation runs. The different scenarios (LD25, HD25, HD100) are sorted by columns from left to right. Different timeout values ($T_m$) are sorted by rows in ascending order.}
    \label{fig:time}
\end{figure}

Figure~\ref{fig:time} evaluates the median recognition latency $T_c$, defined as the time required for 80\% of the swarm to converge on a quorum detection (i.e., $Q\geq0.8$) in challenging conditions. We consider here all configurations where $Q\geq 0.8$ and the ground truth exceeds the decision threshold by a margin of at least $0.1$ (i.e., $G - \tau \geq 0.1$). From these configurations, we select the one with the smallest value of $G$. Hence, for different values of $\tau$, the difference $G-\tau$ varies due to the discrete sampling we performed. This causes artificial oscillations in the latency $T_c$ shown in the plot, because the closer $\tau$ is to $G$, the slower the QS process is.

The baseline \AN appears rapid in converging to detection. However, this speed is a symptom of structural overconfidence; the unchecked double-counting of local opinions accelerates convergence toward premature and biased decisions. The \ANT protocol generally exhibits longer latencies, as it requires more time to gather diverse spatial samples due to the information inertia related to persistence of old messages in the buffer. However, configuring it with shorter timeouts can artificially increase its speed at the direct expense of its accuracy. The \ANTk protocols show similar latencies as \ANT, confirming that the loss in message diversity does not impact accurate QS. 

For all protocols, latency inversely correlates with the decision threshold $\tau$ and the corresponding ground truth $G$. When the population is very heterogeneous in terms of commitment states (lower $\tau$ and $G$), statistical noise induces frequent opinion fluctuations, prolonging the time required for collective stabilization. Conversely, homogeneous populations naturally suppress these fluctuations. Latency is also impacted by density, with the high-density, small arena scenario HD25 resulting in fastest processes, followed by HD100. 

\subsection{System Resilience}
To quantify protocol resilience, we isolated experimental runs where the fixed ground truth $G$ was close to the decision threshold $\tau$, bounded by $|G - \tau| \leq 0.05$. This boundary represents conditions where the swarm is vulnerable to transient misclassification. We tracked individual quorum estimates and extracted two metrics: the average per-robot error rate $E_r$, and the average time required to recover from an erroneous state $T_r$. To characterise the error recovery dynamics and account for right-censored data---such as instances where the simulation concludes before a robot successfully exits an erroneous state---we employed the Kaplan-Meier (KM) estimator to compute the empirical cumulative distribution function of the recovery times. Building upon this estimation, the mean recovery time $T_r$ was subsequently computed from a Weibull distribution fitted to the empirical distribution across $R=100$ runs.
\begin{figure}[!ht]
    \centering
    \includegraphics[width=1\textwidth]{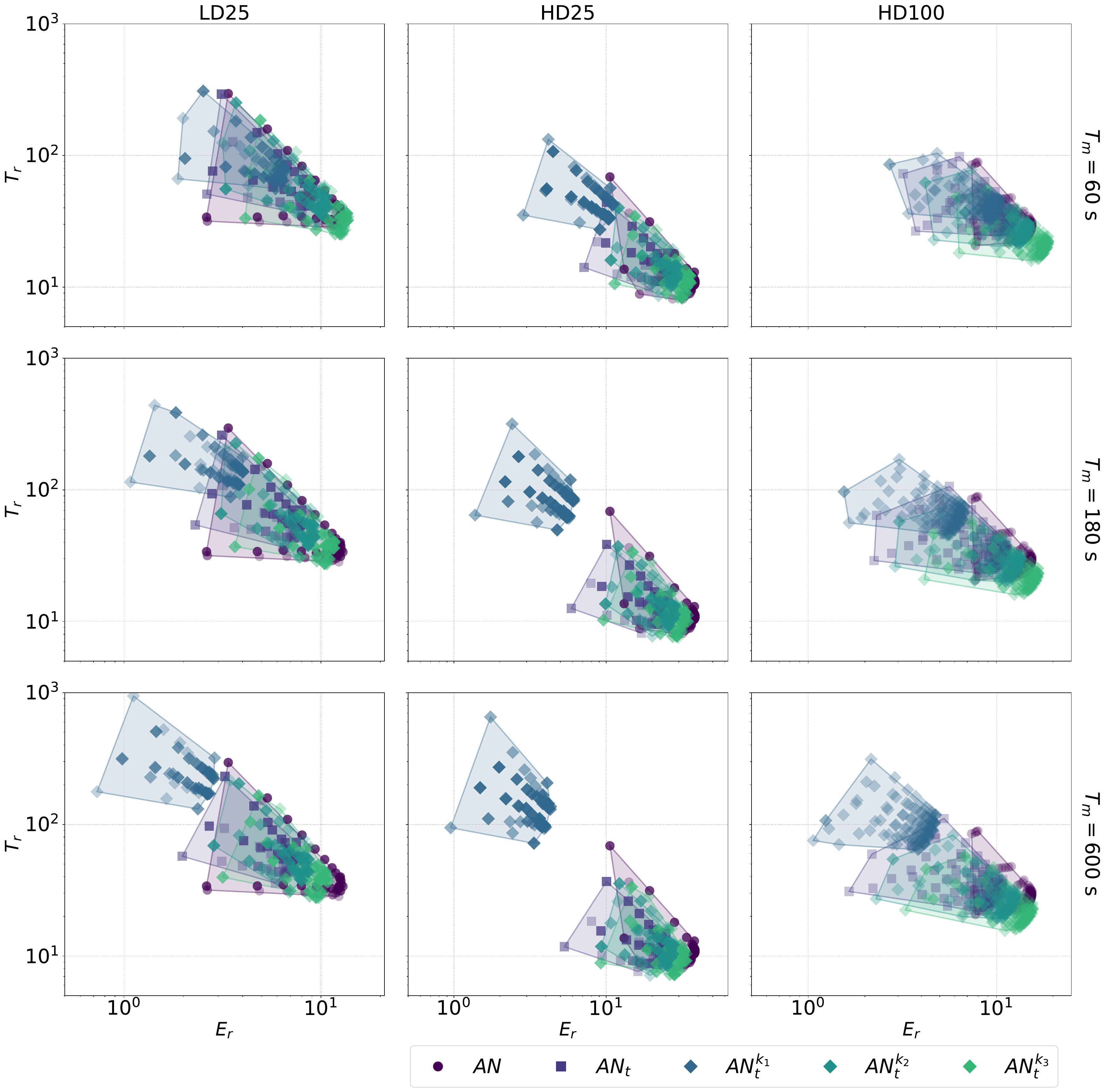}
    \caption{Scatter plot illustrating the trade-off between the average error rate ($E_r$) and the median recovery time ($T_r$) under hard decision conditions ($|G - \tau| \leq 0.05$). The different scenarios (LD25, HD25, HD100) are sorted by columns from left to right. Different timeout values ($T_m$) are sorted by rows in ascending order.}
    \label{fig:scatter_recovery}
\end{figure}

Figure~\ref{fig:scatter_recovery} visualises the relationship between these metrics. The baseline and accuracy-focused protocols (\AN and \ANT) frequently disperse along the $E_r$ axis, exhibiting higher temporary error rates, particularly in dense scenarios where spatial double-counting is pronounced. 
By applying the stability variant protocol (\ANTk), small values of $k$ (e.g., $k=1$) can be utilised to modify the consistency of quorum estimates, especially for high-density conditions and long timeout values. Configuring the system with a minimal filter systematically compresses the error distributions and decreases the average error rate ($E_r$) compared to \ANT. However, this stabilisation implies a clear operational cost: the reduction in temporary errors causes the recovery time ($T_r$) to increase, highlighting a fundamental trade-off between transient responsiveness and steady-state precision. The enhanced stability explains why, despite a lower message diversity, the sampling protocols maintain a good accuracy. Indeed, errors can get reduced consistently in number, however, when they occur, they last longer in general.

\section{Conclusion}\label{sec:conclusions}
This study provides an analysis of QS dynamics under anonymous communication, highlighting the structural vulnerabilities of such approaches and evaluating the efficacy of stochastic filtering. The baseline anonymous protocol \AN, while offering a parsimonious and fast solution, inherently suffers from double-counting. This limitation introduces a severe bias, resulting in structural overconfidence and inaccurate estimations. In real-world robotic deployments, such inaccuracies can be deeply problematic: if just a subset of the swarm successfully detects the quorum threshold, the group risks fragmentation, division of labour failures, or incoherent collective actions.

By introducing two variant protocols, we demonstrated how anonymous message buffers can be managed to improve the QS process. The \ANT protocol mitigates the problem from spatial correlations that results in duplicated messages, improving the message diversity and the  overall accuracy. This comes at the cost of operational speed, suffering from information inertia due to old messages remaining in the buffer for longer times. Meanwhile, the \ANTk protocol implements a filtering technique, inspired by $k$-priority sampling, directly into the message buffers to modulate the stability of the swarm's estimates. 
This stabilisation introduces a distinct operational trade-off within the system's resilience profile: \ANTk ($k=1$) compresses temporary errors ($E_r$) at the expense of a lengthened recovery window ($T_r$) compared to \ANT. Such a trade-off is particularly valuable in high-stakes robotic deployments, such as collective search-and-rescue missions, where robots must reliably detect a structural hazard before triggering an irreversible evacuation or containment protocol. In these scenarios, initiating a critical group response based on a transient false positive carries considerable costs, making the stability and lower error rates of \ANTk far more desirable despite the longer recovery window. Conversely, applications that tolerate minor temporary mistakes in favour of rapid, agile recovery may find the unmasked dynamics of \ANT more suitable. However, a too large value of $k$ triggers ``data starvation'', where useful information is discarded alongside duplicates, thereby losing the stabilisation effect. Hence, exploiting the filtering technique with small values of $k$ is particularly effective when transient individual errors are costly.

Ultimately, these findings demonstrate that anonymous communication buffers can be actively managed to balance accuracy, steady-state decision stability, and transient recovery dynamics. Although stochastic filtering cannot circumvent the absolute bounds on accuracy dictated by the lack of unique identities, it provides a tunable protocol to tailor the reliability of QS in minimalist robot swarms.

\begin{credits}
\subsubsection{\ackname} This work has been supported by the Horizon Europe, PathFinder Open European Innovation Council Work Programme for the project BABOTS under grant agreement No 101098722, by the Italian Ministry of Uninversity and Research MUR for the PNRR project FAIR (PE0000013-FAIR) and by the DFG under Germany's Excellence Strategy, EXC 2117-422037984.

\subsubsection{\discintname} The authors declare that they have no conflicts of interest.
\end{credits}

\bibliographystyle{splncs04}
\bibliography{KpriorityBib}

\end{document}